\documentclass[10pt,twocolumn,letterpaper]{article}

\usepackage{cvpr}
\usepackage{times}
\usepackage{epsfig}
\usepackage{graphicx}
\graphicspath{{figures/}{Figures/}}
\usepackage{amsmath,wrapfig, graphicx,bm}
\usepackage{amssymb,amsthm}
\usepackage{paralist}

\DeclareMathOperator*{\argmin}{arg\,min}

\usepackage{multirow}
\usepackage{colortbl, array}
\usepackage{hhline}
\usepackage[svgnames]{xcolor}
\usepackage{empheq}

\usepackage{booktabs}

\usepackage[utf8]{inputenc} 
\usepackage[T1]{fontenc}    
\usepackage{amsmath,amsthm,amssymb,comment,tabularx}       
\usepackage{url}            
\usepackage{booktabs}       
\usepackage{amsfonts}       
\usepackage{nicefrac}       
\usepackage{microtype}      
\usepackage[linesnumbered,ruled]{algorithm2e}
\usepackage{graphicx,caption,paralist,color,wrapfig}
\usepackage{accents}
\newtheorem{definition}{Definition}
\newtheorem{theorem}{Theorem}

\usepackage{algpseudocode}

\definecolor{rulecolor}{RGB}{70,10,171}
\definecolor{tableheadcolor}{RGB}{120,50,200}

\usepackage[pagebackref=true,breaklinks=true,letterpaper=true,colorlinks,bookmarks=false]{hyperref}

\def\logm{\operatorname{log}}

\cvprfinalcopy
\def\cvprPaperID{1}

\ifcvprfinal\fi
\begin{document}

\title{
\mbox{\hspace{-10mm}C-SURE: Shrinkage Estimator and Prototype Classifier for Complex-Valued Deep Learning}
}
\author{
\setlength{\tabcolsep}{6mm}
\begin{tabular}{@{}cccc@{}}
Rudrasis Chakraborty$^{1}$%
\thanks{Authors of equal contributions.}
&
Yifei Xing$^{1*}$&
Minxuan Duan$^2$&
Stella X. Yu$^1$\\
\end{tabular}\\[5pt]
\setlength{\tabcolsep}{6mm}
\begin{tabular}{cc}
$^1$ UC Berkeley / ICSI&
$^2$  Peking University\\
\end{tabular}\\
}

\maketitle
\begin{abstract}\label{conc}
The James-Stein (JS) shrinkage estimator is a biased estimator that captures the mean of Gaussian random vectors.  While it has a desirable statistical property of dominance over the maximum likelihood estimator (MLE) in terms of mean squared error (MSE), not much progress has been made on extending the estimator onto manifold-valued data.  

We propose C-SURE,  a novel Stein’s unbiased risk estimate (SURE) of the JS  estimator  on the manifold of complex-valued data  with a theoretically proven optimum over MLE.  
Adapting the architecture of the complex-valued SurReal classifier,
we further incorporate C-SURE into a  prototype convolutional neural network (CNN) classifier.

We compare C-SURE with SurReal and a real-valued baseline on complex-valued MSTAR and RadioML datasets. 
C-SURE is more accurate and robust than SurReal, and the shrinkage estimator is always better than MLE for the same prototype classifier.  Like SurReal, C-SURE is much smaller, outperforming the real-valued baseline on MSTAR (RadioML) with less than $1\%$ ($3\%$) of the baseline size.
\end{abstract}
\def\figoverview#1{
\begin{figure*}[#1]
    \centering
    \includegraphics[trim=20 250 0 350, width=0.99\textwidth]{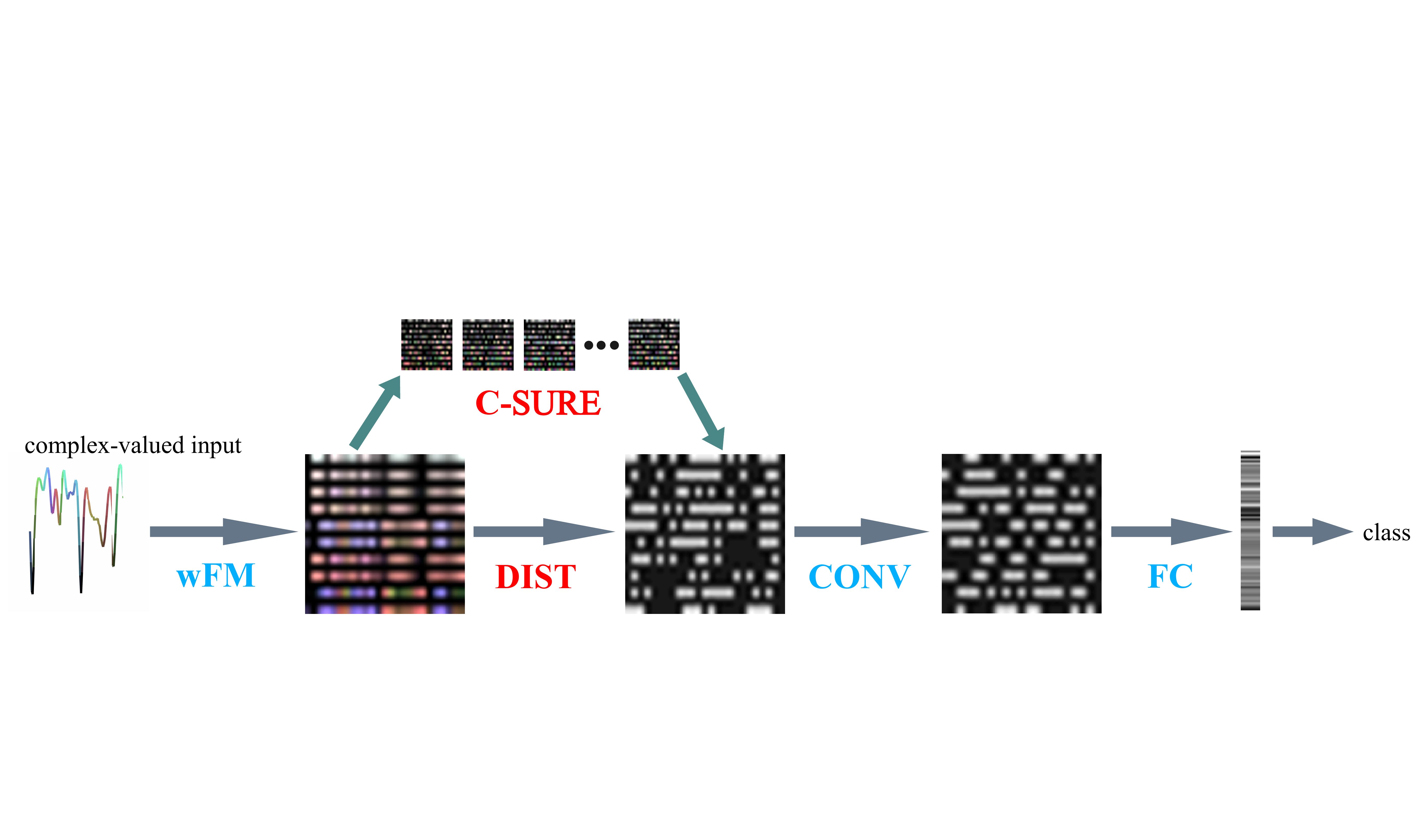}
    \caption{Workflow of our model-based C-SURE CNN classifier.  Our model is based on the SurReal complex-valued CNN \cite{chakraborty2019real}, which consists of wFM convolution layers (wFM), distance transformation layers (DIST), standard convolution layers (CONV), and finally fully connected layers (FC) for softmax classification.  We incorporate C-SURE into the distance transformation layer: During training,  the statistical mean of the wFM features per class is estimated using C-SURE, and the minimum distances between the wFM features and the set of class means become the real-valued output; during testing, only the distances between the wFM features and the saved class means need to be calculated.  The real-valued distances go through standard CONV and FC layers for the final classification into semantic categories. }
    \label{overview}
\end{figure*}
}

\section{Introduction}

Deep learning has been widely adopted in computer vision, often assuming data that  follow vector-space properties.  However, there are plenty of natural non-Euclidean manifold-valued data.  Complex-valued data such as medical images, radio signals, and nuclear covariances can all be modeled as Riemannian manifolds.  Even for real-valued signals, their manifold-valued representations could be more informative of underlying signals, such as Fourier transforms and spectrum-based techniques \cite{freeman1991design, maire2016affinity, yu2011angular}. 

The earliest relevant research for manifold-valued deep learning can be traced back to \cite{sochen1998general}, which regards images as manifolds and applies differential geometry for structural analysis.  More recent works have explored preserving the inherent geometry of graphs \cite{gori2005new, scarselli2008graph}, or achieving group equivariance and invariance \cite{bruna2013spectral,cohen2016group}.  However, these works do not offer an extension to naturally manifold-valued data. 

The intrinsic geometric structures of non-Euclidean manifold data are addressed in \cite{chakraborty2018h, esteves2018polar, chakraborty2018statistical}.  In particular, with convolution on the manifold defined as weighted Fr\'{e}chet mean (wFM) filtering \cite{chakraborty2018h, chakraborty2018manifoldnet}, significant performance gain can be achieved on manifold-valued data along with drastic reduction in the model parameter count.

We focus on the manifold of complex-valued data such as synthetic  aperture  radar  (SAR)  images,  magnetic resonance  (MR)  images,  and radio  frequency  (RF)  signals.   Per the polar form of complex numbers, the complex plane can be treated as a product space of scaling and planar rotations.  This representation allows \cite{chakraborty2019real} to develop an efficient CNN classifier based on wFM filtering on the specific manifold.

The wFM is only one way to compute the mean of samples on a manifold.  Recently, \cite{yang2019shrinkage} suggests an alternative by defining a James-Stein (JS) estimator that outperforms the Fr\'{e}chet mean in terms of MSE over the field of semi-positive definite matrices (SPD).  Here we extend the idea to data lying on the field of complex numbers and prove the dominance of the  JS estimator over the Fr\'{e}chet mean. 

The JS estimator arises from simultaneously estimating the mean of a multivariate homoscedastic normal distribution.  Let $X$ denote a random vector whose $p$ components are independent and normally distributed with mean $\theta_i$ and variance $\sigma^{2}$, $i=1,\ldots,p$.  When $p>2$, it can be shown that the JS estimator $\theta^{\operatorname{JS}}$ dominates the maximum likelihood estimator (MLE) 
$\theta^{\operatorname{MLE}}$ 
\cite{james1992estimation}:
\begin{align}
\theta^{\operatorname{MLE}}&= X\\
\theta^{\operatorname{JS}}&=\left(1-\frac{(p-2) \sigma^{2}}{\|X\|^{2}}\right) X.
\end{align}
When $(p-2)\sigma^2<\|X\|^2$, the JS estimator simply takes the natural estimator $X$ (i.e., $\theta^{\operatorname{MLE}}$) and shrinks it towards the origin 0.   The JS estimator can be viewed as an empirical Bayes method, where $\theta$ itself is a random variable with prior distribution and needs to be estimated \cite{efron1973stein}.  
\figoverview{tp}

The JS estimator can be viewed as a special case of a hierarchical Bayesian model \cite{berger1996choice}, where the unknown mean $\theta_i$ follows a normal prior distribution with mean $\mu$ and variance $\tau^2$.  Given data variance $\sigma^2$, for any prior setting $(\mu,\tau)$, the maximum a posterior (MAP) estimation of $\theta$ is a weighted sum of the data and the prior:
\begin{align}
\hat{\theta}(\mu,\tau;\sigma)=\frac{\tau^2}{\tau^2+\sigma^2} X+\frac{\sigma^2}{\tau^2+\sigma^2} \mu.
\label{JSMAP}
\end{align}
The best JS estimator can be solved by choosing   $(\mu,\tau)$ that minimizes the Stein’s unbiased risk estimate (SURE) \cite{stein1981estimation}:
\begin{align}
&\hat{\mu}^{\text{SURE}}, \hat{\tau}^{\text{SURE}}=\argmin _{\mu, \tau} \operatorname{SURE}(\mu, \tau)    
\\
&\operatorname{SURE}(\mu, \tau)=-p \sigma^{2}+\|\hat{\theta}-X\|^{2}+2 \sigma^{2} \sum_{i=1}^{p} \frac{\partial \hat{\theta}}{\partial X_{i}}
\label{SURE}
\end{align}
The importance of SURE is that it does not depend on the unknown $\theta$, and yet it is an unbiased estimate of the mean-squared error (MSE) between $\hat{\theta}$ and $\theta$.  Minimizing SURE can thus act as a surrogate for minimizing the MSE and the optimal estimation setting can be obtained without $\theta$.

It has been a challenge to generalize the JS shrinkage estimator to non-Euclidean spaces. 
The idea of shrinking is natural only on certain manifolds; there is no formula for the shrinkage estimator on manifolds in general.  For instance, a shrinkage estimator for covariance matrices is designed in \cite{ledoit2004well,daniels2001shrinkage} and then generalized to the Riemannian manifold of symmetric positive definite (SPD) matrices \cite{yang2019shrinkage}.  

Our goal is not only to extend the shrinkage estimator to the manifold of complex numbers, but also to use it in learning the classification of signals or images, an under-explored application of the JS estimator among various machine learning settings
\cite{manton1998james,wu2013james,fleishman2015geodesic}.

We propose C-SURE, a novel SURE of the JS estimator on the manifold of complex-valued data with a theoretically proven optimum over MLE.  We incorporate it into learning a convolutional neural network (CNN) classifier (Fig.\ref{overview}).  Instead of learning a purely discriminative classifier, we learn a nearest prototype-based classifier based on the feature distribution mean of each class. 

Compared to popular real-valued CNN classifiers and the SurReal CNN classifier \cite{chakraborty2019real} based on the Fr\'{e}chet mean on the complex-valued manifold, our C-SURE prototype classifiers achieve better performance with faster convergence on two complex-valued datasets: MSTAR and RadioML. 
Our model is also much smaller, outperforming the real-valued CNN on MSTAR with less than $1\%$ of the model size.

\def\figAlg#1{
\begin{algorithm}[#1]
\SetKwInput{KwData}{Input}
\SetKwInput{KwResult}{Output}
 \KwData{data \(X_{\text{all}}\), variances  \(\{v\}\)}
 \KwResult{class means \(\{{M_i}\}\), distance features \(\{{O_i}\}\),\\
 \hfill $i$ out of $p$ refers to the $i$-th dimension of the data.
 }
\For{each class}{
 Gather instances of this class in $X$\\
 Calculate a running estimate of \(\overline{{X}}^{\mathrm{LE}}\)\\
   \If{training}{
   \For{each mixture component $k$}{
\scalebox{1}{$
\begin{array}{r}\displaystyle
\left(\hat{\mu}_k^{\text{SURE}},\hat{\lambda}_k^{\text{SURE}}\right)=\argmin_{ \mu_k,  \lambda_k} \sum_{i=1}^{p} \frac{v}{\left(\lambda_k+v\right)^{2}}\\
\left(v\left\|\log \overline{X}_{i}^{\mathrm{LE}}-\log \mu_k\right\|^{2} +\frac{p\left(\lambda^2_k-v^2\right)}{N}\right)
\end{array}
$
}}}
  Calculate the C-SURE estimate \(\) according to Eqn.(\ref{estimator})\\
\scalebox{0.72}{$\displaystyle
\widehat{M}_i^{\mathrm{SURE}}(w)=\sum_{k=1}^K \exp \left(w_k\left( \frac{\hat{\lambda}_k^{\text{SURE}}}{\hat{\lambda}_k^{\text{SURE}}+v} \log \overline{{X}}_{i}^{\mathrm{LE}}+\frac{v}{\hat{\lambda}_k^{\text{SURE}}+v} \log \hat{\mu}_k^{\text{SURE}}\right)\right)
$}\\
Compute \(d\left(\widehat{M}_i^{\mathrm{SURE}}(w), \overline{X}_{i}^{\mathrm{LE}}\right)\)}
Compute $O_i$ as the minimal distance between $X_i$ and all the class means \(\left\{\widehat{M}_{i}^{\mathrm{SURE}}(w)\right\}\)
\\
  Update $w$ with SGD, to reduce the classification loss
 \caption{C-SURE Prototype Feature Layer.}
 \label{fig:alg}
\end{algorithm}

}

\section{Shrinkage Estimator of Complex Numbers}
We view the field of complex numbers $\mathbf{C}$ as a product group of two smooth Riemannian manifolds, or more specifically Lie groups.  On each of these two manifolds, we define Gaussian distributions and construct the JS shrinkage estimator of the Fr\'{e}chet Mean (FM) on the manifold.  We show that our shrinkage estimator on $\mathbf{C}$ yields a uniformly smaller risk than the MLE estimator, i.e., FM.
\\

\noindent
{\bf Manifold View of Complex Plane $\mathbf{C}$.}  
We represent $\mathbf{C}$ as a product space of two Riemannian manifolds \cite{chakraborty2019real}.  Utilizing the polar form of a complex number $\mathbf{c} = r{e^{i\theta}}$, we have:
\begin{align}
\mathbf{C}  = \left\{re^{i\theta}\right\}\simeq \mathbf{R^+}\times \mathbf{S^1}\simeq \mathbf{P_1} \times \textsf{SO(2)} 
\end{align}
where the manifold of $1\times 1$ semi-positive definite matrices $\mathbf{P_1}$ is topologically the space of non-negative numbers $\mathbf{R^+}$, and the manifold of $2\times 2$ rotation matrices $\textsf{SO(2)}$ is topologically a circle $\mathbf{S^1}$.  With this decomposition, designing the JS shrinkage estimator on  $\mathbf{C}$ is reduced to designing the  estimator on  $\mathbf{P_1}$ and $\textsf{SO(2)}$ separately.

The shrinkage estimator on  $\mathbf{P_1}$ has been dealt with in \cite{yang2019shrinkage}, where they propose a novel shrinkage estimator on the FM of SPD matrices, with proven dominance over the MLE of the FM in terms of the MSE risk.  

In order to develop the shrinkage estimator on $\textsf{SO(2)}$, we choose a Riemannian metric and define our FM.  Using the Lie algebra of $\textsf{SO(2)}$, we can apply the procedure in \cite{yang2019shrinkage} and derive our  shrinkage estimator on $\textsf{SO(2)}$.
\\[1pt]

\noindent
{\bf Fr\'{e}chet Mean on a Riemannian manifold.}
Let $\mathcal{M}$ be a topological manifold equipped with a Riemannian metric, and let $d:\mathcal{M}\times \mathcal{M}\rightarrow \mathbf{R}$ denote the associated distance.  Given a collection of $n$ points $\{X_{i},i=1,\ldots,n\}$ on the manifold, their Fr\'{e}chet Mean  \cite{Frechet1948elements} is defined as:
\begin{equation}
\bar{X}=\underset{X\in \mathcal{M}}{\argmin } \sum_{i=1}^{n} d^{2}(X, X_{i}).
\end{equation}
In general, $\bar{X}$ may not be unique, but can be made unique under certain constraints.\\

\noindent
{\bf Log-Euclidean Metric on $\textsf{SO(2)}$.}
We endow $\textsf{SO}(2)$ with the Log-Euclidean (LE) metric, for it is computationally efficient with closed-form solutions.  We extend the Log-Euclidean metric in  \cite{arsigny2007geometric} and define the induced  geodesic distance \(d_{L E} : \textsf{SO(2)} \times \textsf{SO(2)} \rightarrow \mathbf{R}\) as:
\begin{align}
&d_{LE}(X_1, X_2) =\|\logm(X_1)-\logm(X_2)\|_{F}\\
&X_{i}  =\left[\begin{array}{rr}{\cos \theta_{i}} & {\!-\!\sin \theta_{i}} \\ {\sin \theta_{i}} & {\cos \theta_{i}}\end{array}\right]\in  \textsf{SO(2)}\\
&\logm(X_i) =(\theta_i\!+\!2\pi k)\!
\left[\setlength{\arraycolsep}{2pt}\begin{array}{rr}{0} & {-1} \\ {1} & {0}\end{array}\right],\, k=\pm 1,\pm 2,\ldots.
\end{align}
Here $\logm$ denotes the matrix logarithm.  The logarithm of a rotation matrix is not unique; we fix $k=0$ here and obtain an isomorphism between $\mathfrak{so}(2)$ and the interval $(-\pi,\pi]$. This particular logarithm is called \textit{principal}.

We establish a mapping $\widetilde{X}$ from point $X$ on the manifold 
$\cal{M}\!=$\textsf{SO(2)} to a number in the real domain $\bf{R}$ that indicates  the size of rotation directly.  Since the Lie algebra $\mathfrak{so}(2)$ of  \textsf{SO(2)} is the space of $2\!\times\!2$ skew-symmetric matrices, 
we define $\widetilde{X}$ through the mapping $\Phi: \mathfrak{so}(2)\to \bf{R}$.
\begin{align}
\widetilde{X}&=\Phi(\logm(X))  =\sqrt{2}\,\theta\\
\logm(X) & =\left[\begin{array}{cc}{0} & {-\theta} \\ {\theta} & {0}\end{array}\right]\in\mathfrak{so}(2).
\end{align}
The distance between points $X_1$ and $X_2$ on \textsf{SO(2)} is simply:
\begin{align}
\hspace{-2mm}
d_{L E}(X_1,X_2)\!=\!\min\{|\widetilde{X_1}\!-\!\widetilde{X_2}|,2\sqrt{2}\pi\!-\!|\widetilde{X_1}\!-\!\widetilde{X_2}|\},
\end{align}
ensuring the 
uniqueness of the FM on \textsf{SO(2)}.\\

\noindent
{\bf Gaussian Distributions on \textsf{SO(2)}.}
Gaussian distributions on the manifold of positive definite matrices become Log-Normal distributions \cite{schwartzman2006random}.  We follow the same procedure and extend it to other matrix Lie groups such as $\textsf{SO(2)}$.
 \begin{definition}
We say $X$ follows a Log-Normal distribution with mean $M$ and covariance matrix $\Sigma \in \mathbf{R}_{m\times m}$, or \(X \sim \operatorname{LN}(M, \Sigma)\) if
\begin{equation}
    \widetilde{X} \sim N(\widetilde{M}, \Sigma)
\end{equation}
\end{definition}
We can further define the mixture of Gaussians in order to capture multi-modal distributions in real-world data. 
\begin{definition}
We say that $X$ follows a mixture of $K$ Log-Normal distributions each with mean $M_k \in G$ and covariance matrix $\Sigma_k \in \mathbf{R}^{m\times m}$, or \(X \sim \operatorname{MLN}(w , \boldsymbol{M}, \boldsymbol{\Sigma})\) if
\begin{align}
& \widetilde{X} \sim \sum_{k=1}^{K}w_{k}N(\widetilde{M}_k, \Sigma_k)\\
& \sum_{\mathrm{k}=1}^{\mathrm{K}} w_{\mathrm{k}}=1,
\quad 0 \leq w_{\mathrm{k}} \leq 1, \forall k.
\end{align}
\end{definition}
We refer to each $N(\widetilde{M}_k, \Sigma_k)$ as the $k$-th component density in the Gaussian mixture model.  
 
Calculating these distributions requires the composition of logarithmic and exponential maps.  On $\textsf{SO(2)}$, we have
\begin{align}
\operatorname{exp}\left(\operatorname{log} X + \operatorname{log} Y\right)=XY
\end{align}
since $\textsf{SO(2)}$ has a trivial Lie algebra with zero Lie brackets. 
\\

\noindent
{\bf C-SURE Shrinkage Estimator on $\text{\sf SO(2)}^p$.} 
Let $X$ denote a $p$-dimensional complex-valued random variable; for the $i$-th dimension, the value $X_i$ is modeled as a point on the manifold $\textsf{SO(2)}$.  We are going to estimate the $p$-dimensional mean vector $M$ from a collection of these manifold-valued observations, using the JS estimator derived from a hierarchical Bayesian approach.

For $X\in\textsf{SO(2)}^p$, we assume that $X_i$ is independently distributed according to the Log-Normal with individual mean $M_i$ and equal variance $vI$, and the means \{$M_i,i=1,\!\ldots,\!p$\} are independently and identically distributed according to a Log-Normal mixture:
\begin{align}
 X_{i} | M_{i} & \stackrel{ind}{\sim} \operatorname{LN}\left(M_{i}, v I\right), i=1, \ldots, p\\ 
      M_{i} &\stackrel{i.i.d}{\sim} \operatorname{MLN}({w} ,{\mu}, {D}).
\end{align}
We assume that $v$ is known and ${w}$ is fixed, whereas ${\mu}$ and ${D}=\operatorname{Diag}(\lambda_1 I, \cdots , \lambda_K I)$ are unknown and can be optimized by minimizing the SURE risk.  


Since $w$ is fixed, we can first calculate the JS estimator for each of the $k$ component densities independently, and 
then combine them with their respective weights in ${w}$ to obtain the JS estimator of the Log-Normal mixture.

Specifically, using the derivations for the Gaussian distribution on \textsf{SO(2)}, we  extend the MAP estimate in Eqn(\ref{JSMAP}) to \textsf{SO(2)} for the $k$-th component density \(M_{i,k}\sim \operatorname{LN}\left(\mu_{k}, D_k\right)\):
\begin{align}
   \widehat{M}_{i,k}^{ {\mu},  {D}}\left( {w}\right)=\exp \left(\frac{\lambda_k}{\lambda_k\!+\!v} \log \overline{X}_{i}^{\mathrm{LE}}({w})\!+\!\frac{v}{\lambda_k\!+\!v} \log \mu_k\right)
\end{align}
where $\overline{X}_{i}^{\mathrm{LE}}({w})$ denotes the mean of $X_i$ over a total of $N$ sample observations, according to the Log-Euclidean metric and given the mixture weights ${w}$.

We can then extend the MSE to our manifold by defining the empirical loss function $l$ as:
\begin{align}
    l\left(\widehat{M}_k^{ \mu,  D}, {M_k}\right)= \sum_{i=1}^p d_{\mathrm{LE}}^{2}
    \left(\widehat{M}_{i,k}^{ \mu,  D},  M_{i,k}\right)
\end{align}
and the corresponding risk $R$ as $\mathbb{E}[l]$:
\begin{align} 
&R\left(\widehat{M}_k^{ {\mu}, {D}}, M_k\right) =\mathbb{E}\left[  l\left(\widehat{M}_k^{ {\mu},  {D}}, {M_k}\right) \right]\nonumber\\
&=\sum_{i=1}^{p} \frac{v}{\left(\lambda_k+v\right)^{2}}\left(v\left\|\log \mu_k-\log M_{i,k}\right\|^{2}+\frac{p\lambda_k^2}{N}\right).
\end{align}  
The SURE estimate in Eqn(\ref{SURE}), $\operatorname{SURE}( \mu_k,  \lambda_k)$, becomes:
\begin{align}
\sum_{i=1}^{p} \frac{v}{\left(\lambda_k\!+\!v\right)^{2}}\left(v\left\|\log \overline{X}_{i}^{\mathrm{LE}}\!-\!\log \mu_k\right\|^{2}\!+\!\frac{p\left(\lambda^2_k\!-\!v^2\right)}{N}\right).
\end{align}
Our SURE estimate of the $k$-th component mean and variance on the manifold of complex values is thus:
\begin{align} 
&\hat{\mu}_k^{\text{SURE}},\hat{\lambda}_k^{\text{SURE}}=\argmin _{ \mu_k,  \lambda_k} \operatorname{SURE}( \mu_k,  \lambda_k) 
\\ 
&=\argmin _{ \mu_k,  \lambda_k} \sum_{i=1}^{p} \frac{v}{\left(\lambda_k+v\right)^{2}}
\nonumber\\
&\left(v\left\|\log \overline{X}_{i}^{\mathrm{LE}}-\log \mu_k\right\|^{2} +\frac{p\left(\lambda^2_k-v^2\right)}{N}\right).
\label{kSURE}
\end{align}
We propose our C-SURE shrinkage estimator for $M_i$ as a weighted sum of its components: 
\begin{align}
&\widehat{M}_{i}^{\text{SURE}}\left(w\right)=
\nonumber\\
&\hspace{-10pt}
\sum_{k=1}^K \exp \left(w_k\left( \frac{\hat{\lambda}_k^{\text{SURE}}}{\hat{\lambda}_k^{\text{SURE}}+v} \log \overline{{X}}_{i}^{\mathrm{LE}}+\frac{v}{\hat{\lambda}_k^{\text{SURE}}+v} \log \hat{\mu}_k^{\text{SURE}}\right)\right).
\label{estimator}
\end{align}

\noindent
{\bf Optimality of C-SURE Shrinkage over MLE.}
We show that \(\left(\hat{\mu}_k^{\text { SURE }}, \hat{\lambda}_k^{\text { SURE }}\right)\)  minimizes  the actual risk \(R\left(\widehat{M}_k^{ \mu,  D}, {M_k}\right)\).  We follow the approach in \cite{yang2019shrinkage}: For each component density, we have \(M_{i} \sim \operatorname{LN}\left(\mu, D\right)\) 
where $D=\lambda I$, SURE\((\mu, \lambda)\) is a good approximation of \(l\left(\widehat{M}^{\mu, D}, {M}\right)\).

\begin{theorem}
Assume that
\begin{flushleft}
$\begin{array}{l}{(A) \text{ } v^2<\infty} \\ {(B) \limsup _{ p \rightarrow \infty} \frac{1}{p} \sum_{i} \left\|\log M_{i}\right\|^{2}<\infty} \\ {(C) \limsup _{ p\rightarrow \infty} \frac{1}{ p} \sum_{i}\left\|\log M_{i}\right\|^{2+\delta}<\infty \text { for some } \delta>0}\end{array}$.
\end{flushleft}
Then the following holds in probability as $ p\rightarrow \infty$:
\begin{equation}
\begin{aligned}
\sup _{\substack{\lambda>0\\ \|\log \mu\|<\max _{i}\left\|\log \overline{{X}}_{i}^{L E}\right\|}}\left|\text{SURE}( \mu, \lambda)\!-\!l\left(\widehat{{M}}^{\mu, D}, {M}\right)\right| \rightarrow 0.
\end{aligned}
\end{equation}
\end{theorem}

We can now show that for each component density, our proposed shrinkage estimator is asymptotically optimal, compared with MLE of the FM of Log-Normal distribution on $\textsf{SO(2)}$ in terms of risk. 
\begin{theorem}
If (A), (B), (C) in Theorem 1 hold, then,
\begin{equation}
    \lim _{ p \rightarrow \infty}\left[R\left(\widehat{{M}}^{\text{SURE}}, {M}\right)-R\left(\widehat{{M}}^{\mu,D}, {M}\right)\right] \leq 0
\end{equation}
\end{theorem}

A proof of the theorems above for SPD can be found in \cite{yang2019shrinkage}.  Since the key is the trivial Lie algebra, which holds for both SPD and $\textsf{SO(2)}$, we omit the similar proof.
\\

\noindent
{\bf Weight Update.}
After we obtain the class-wise means \(\widehat{{M}}^{\mathrm{SURE}}\left( w\right)\) using Eqn(\ref{estimator}) for a fixed $w$, we update $w$ by some learning method.  While there are a plethora of statistical algorithms readily available to update $w$, e.g., Bayesian methods, EM algorithm etc., we find that gradient descent is more stable and produces more optimized values. 

\section{Prototype Classifier with C-SURE}

We incorporate our C-SURE shrinkage estimator into a nearest prototype CNN classifier.  We model each class as a mixture of Gaussians and learn their prototypes using our C-SURE estimator.  
Instead of assigning an instance to the nearest prototype, we use their minimal distance as a feature for discriminative classification. 
\\

\noindent
{\bf C-SURE Classifier Architecture.} 
Specifically, we build our classifier based on the SurReal complex-valued CNN \cite{chakraborty2019real}, which consists of wFM convolution layers (wFM), distance transformation layers (DIST), and standard convolution layers (CONV), and finally fully connected layers (FC) for softmax classification (Fig.\ref{overview}). 

We incorporate C-SURE into the distance transformation layer: During training, the statistical mean of the wFM features per class is estimated using C-SURE, and the minimum distances between the wFM features and the set of class means become the real-valued output; during testing, only the distances between the wFM features and the saved class means need to be calculated.  The real-valued distances go through standard CONV and FC layers for the final classification into semantic categories.\\

\figAlg{tp}

\noindent
{\bf C-SURE Prototype Feature Layer.} 
This layer consists two parts, C-SURE and DIST in Fig.\ref{overview}.  The feature from the wFM convolutional layer becomes the input $X$, and it is processed per class as well as per mixture component $k$.  The output is the minimal distance between each instance and the C-SURE estimate of all the class means (Algorithm \ref{fig:alg}).\\

\noindent
{\bf Discussions.} 
We have proposed a novel JS shrinkage estimator on the manifold of complex values and used the SURE estimate to compute the FM on the manifold.  This method is used to calculate the class-specific distribution mean on the manifold and implement a prototype-based classifier.

The dominance of the JS estimator over MLE is a statistical property for some fixed data.  When the shrinkage estimator is incorporated into the loop of deep learning for classification, it is unclear whether the shrinkage estimator has any practical advantage over the simpler MLE.

There are two major changes to the fixed data mean estimation setting: {\bf 1)}
Both the data and the estimator are changing during learning; {\bf 2)} The final task performance is not critically dependent on how well the estimator fits the data mean, but on how well the feature derived from the distance to data prototypes separates classes. 

Therefore, while JS dominates MLE, it is unclear whether a prototype CNN classifier with a built-in JS estimator would be theoretically superior to a purely discriminative classifier such as the SurReal CNN classifier \cite{chakraborty2019real}.

We turn to experiments on complex-valued data classification to test our ideas and validate our approach in practice.

\def\figResSum#1{
\begin{table}[#1]
\centering
\caption{Comparison of Classification Accuracies.
\label{resultSummary}
}
\begin{tabular}{l|c|c|c}
\toprule

\textbf{Dataset} & \textbf{Real-Valued} & \textbf{SurReal} &  \textbf{C-SURE}\\
\midrule

{MSTAR-L} & 99.1\% & \bf 99.2\% & \bf 99.2\% \\
\hline
{MSTAR-S} & 97.4\% & 97.7\% & \bf 98.1\% \\
\hline 
{RadioML} & 75.8\% & 78.4\% & \bf 81.6\% \\
\bottomrule
\end{tabular}
\end{table}
}

\def\figModelSize#1{
\begin{figure}[#1]
\centering
\includegraphics[trim=50 170 70 198,width=0.45\textwidth]{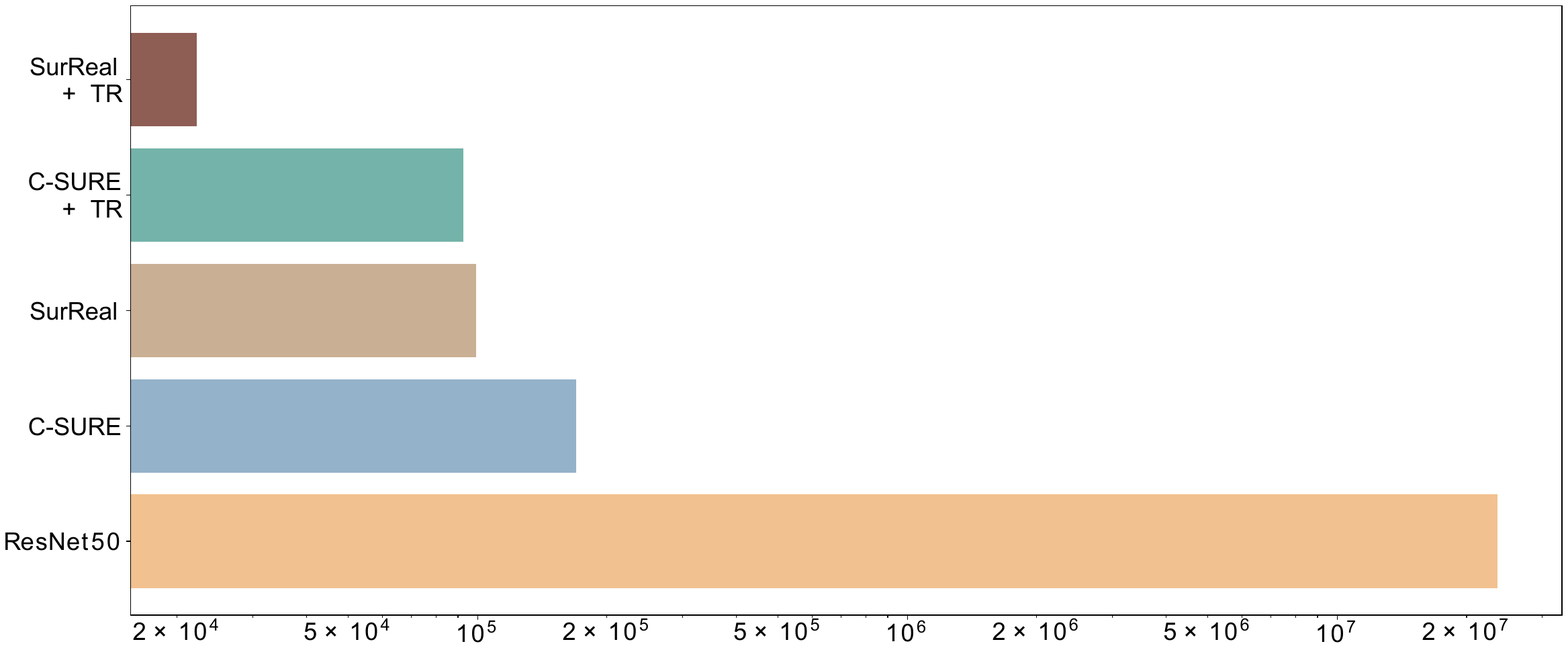}
\caption{MSTAR model size comparison.  Each horizontal bar indicates the total number of parameters on a {\it log} scale.  There are 5 models: two-channel real-valued baseline ResNet50, the complex-valued baseline SurReal, our complex-valued C-SURE, and the latter two models implemented with the tensor ring trick \cite{oseledets2011tensor} and   marked by {$+$TR}.
Our C-SURE classifier has more parameters than the SurReal model that it adapts from.  Like SurReal, C-SURE is smaller than $1\%$ of the real-valued CNN baseline ResNet50.}
    \label{fig:mstar_params}
\end{figure}
}

\def\figMSTARSets#1{
\begin{figure}[#1]
    \centering
    \setlength{\tabcolsep}{1pt}
    \begin{tabular}{cc}
    \includegraphics[trim=40 50 10 50, width=0.24\textwidth]{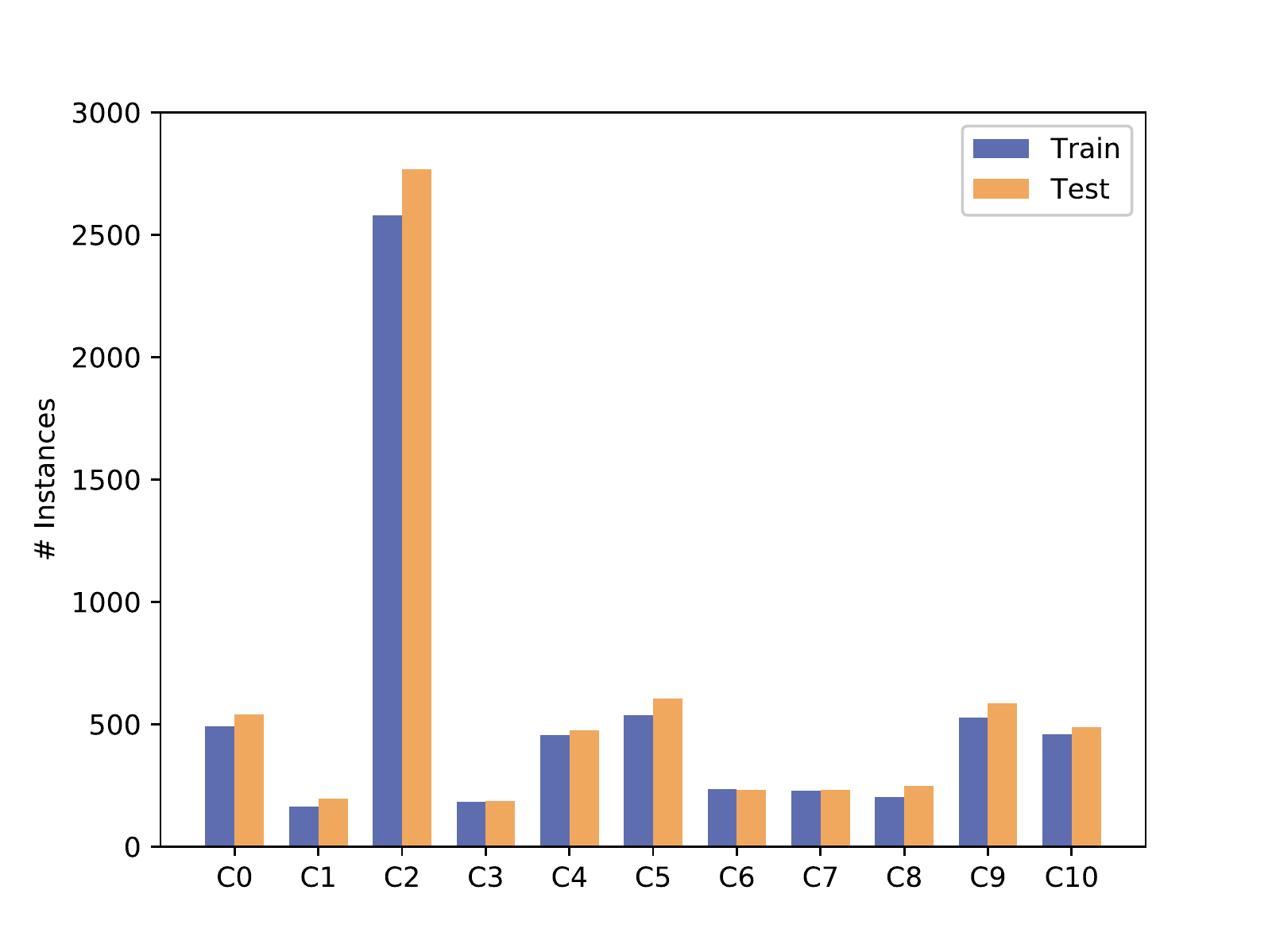} &
    \includegraphics[trim=40 50 10 50, width=0.24\textwidth]{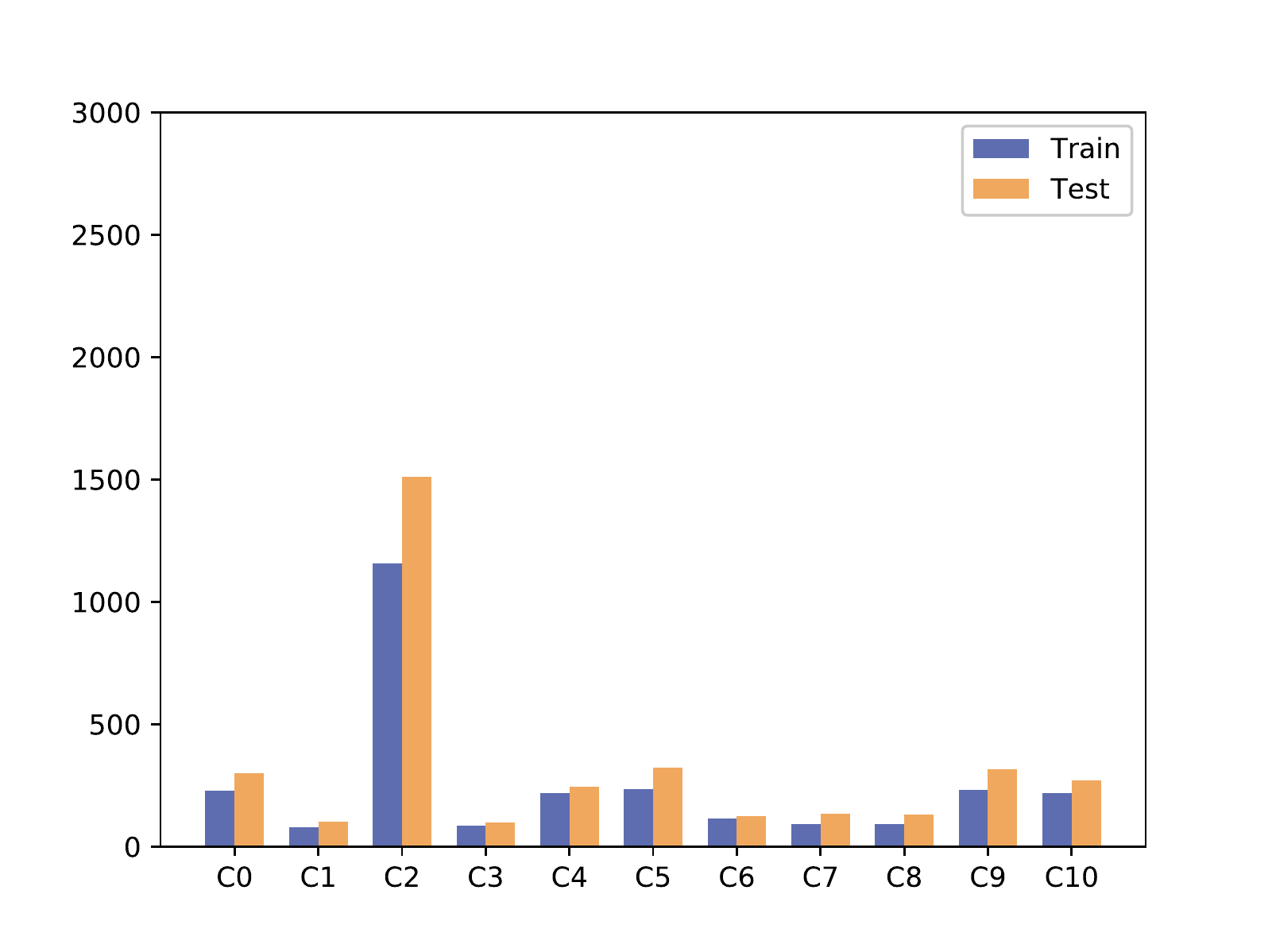} \\
    \end{tabular}
    \caption{MSTAR large (L) and small (S) subsets have highly imbalanced classes.  C2 is the largest class,  C0, C4, C5, C9, C10 come next at about 20\% of the size of C2, and C1, C3, C6, C7, C8 are at about 8\% of the size of C2.}
    \label{fig:mstarSets}
\end{figure}
}

\def\tabSeenSplit#1{
\begin{table*}[#1]
\centering
\caption{\#Instances Per Class in the Unseen Setting}
\label{tab3}
\setlength{\tabcolsep}{3pt}
\begin{tabular}{l|ccc|c|ccc|ccccccc|c}
\toprule
\multirow{2}{0em}{\textbf{Set}} & \multicolumn{3}{ c |}{\textbf{BMP2}}  & \textbf{BTR70} & \multicolumn{3}{ c |}{\textbf{T72}} & \textbf{BTR60} &  \textbf{2S1} & \textbf{BRDM2} & \textbf{D7} & \textbf{T62} & \textbf{ZIL131} & \textbf{ZSU23/4} & \textbf{Clutter} 
\\
\cline{2-16}
& snc21 & sn9563 & sn9566 & c71 & sn132 & sn812 & sns7 & & & & & & & & \\
\midrule

Train & 233 & & & 233 & 232 & & & 256 & 299 & 298 & 299 & 299 & 299 & 299 &  \\ \midrule
Test & 196 & 195 & 196 & 196 & 196 & 195 & 191 & 195 & 274 & 274 & 274 & 273 & 274 & 274 & 1159 \\ 
\bottomrule
\end{tabular}
\end{table*}
}

\def\figAngleConf#1{
\begin{figure}[#1]
    \centering
    \setlength{\tabcolsep}{1pt}
    \begin{tabular}{ccc}
        {\bf a)} ResNet 50 \\
\includegraphics[width=0.36 \textwidth]{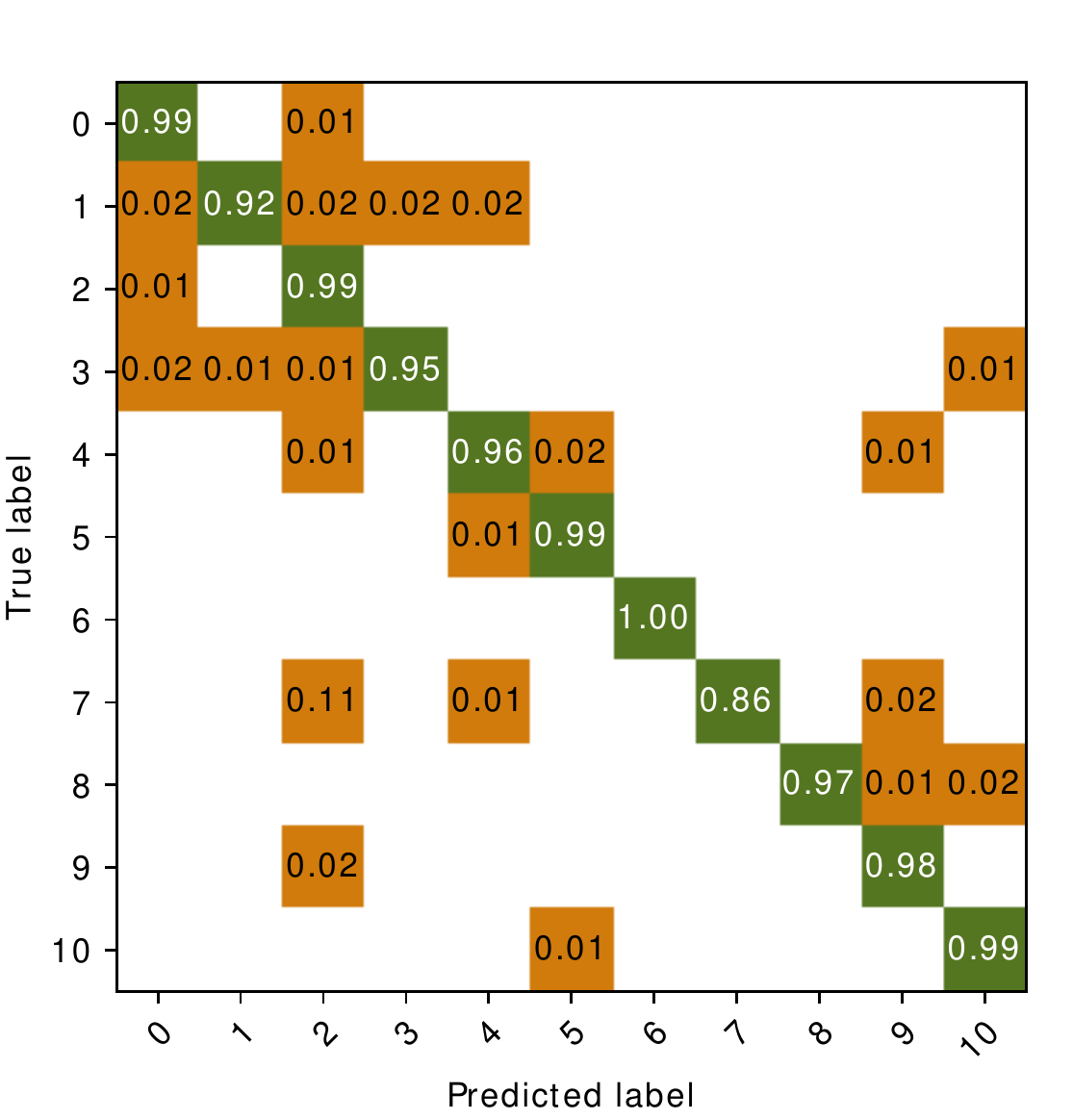}\\
    {\bf b)} SurReal \\
    \includegraphics[width=0.36  \textwidth]{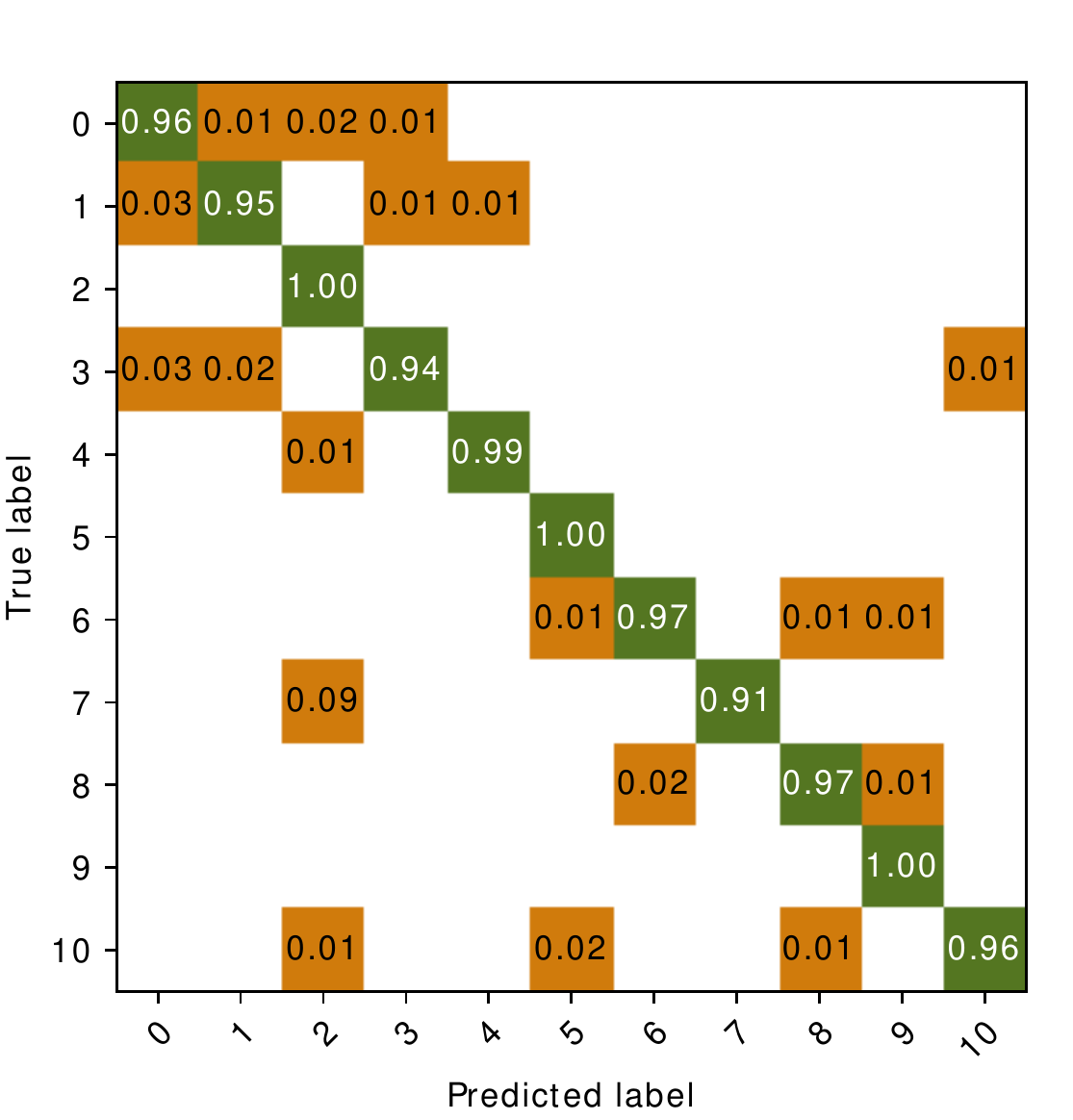}\\
   {\bf c)} C-SURE\\ 
    \includegraphics[width=0.36 \textwidth]{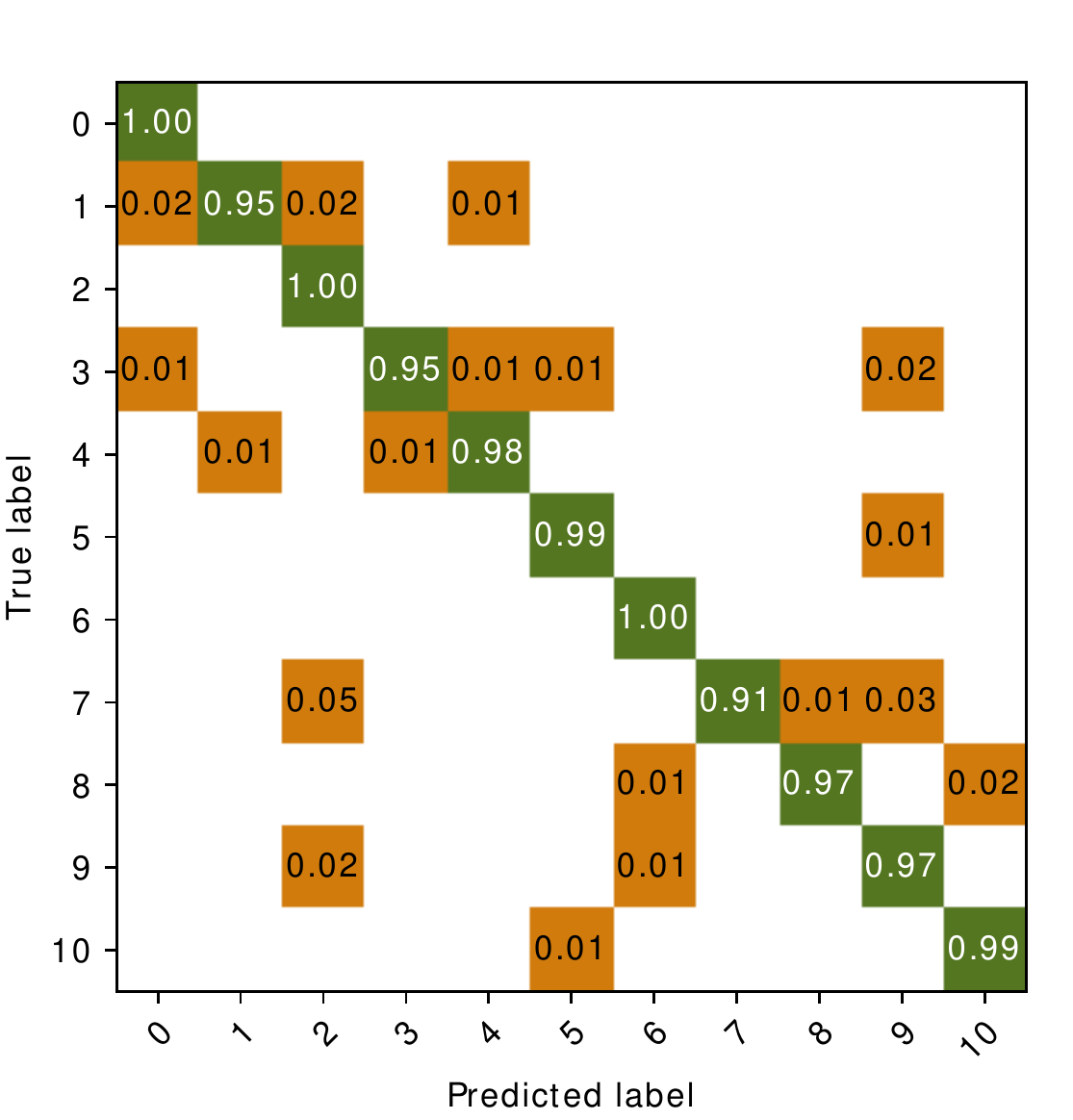}\\
    \end{tabular}
    \caption{C-SURE has the least confusion between classes on our MSTAR small set.  Among the three classifiers,
    the confusion matrix for our C-SURE has the largest values on the diagonal and the smallest values off the diagonal.     \label{fig:conf}
    }
\end{figure}
}

\def\figSeen#1{
\begin{figure}[#1]
    \centering
    \includegraphics[width=0.49\textwidth]{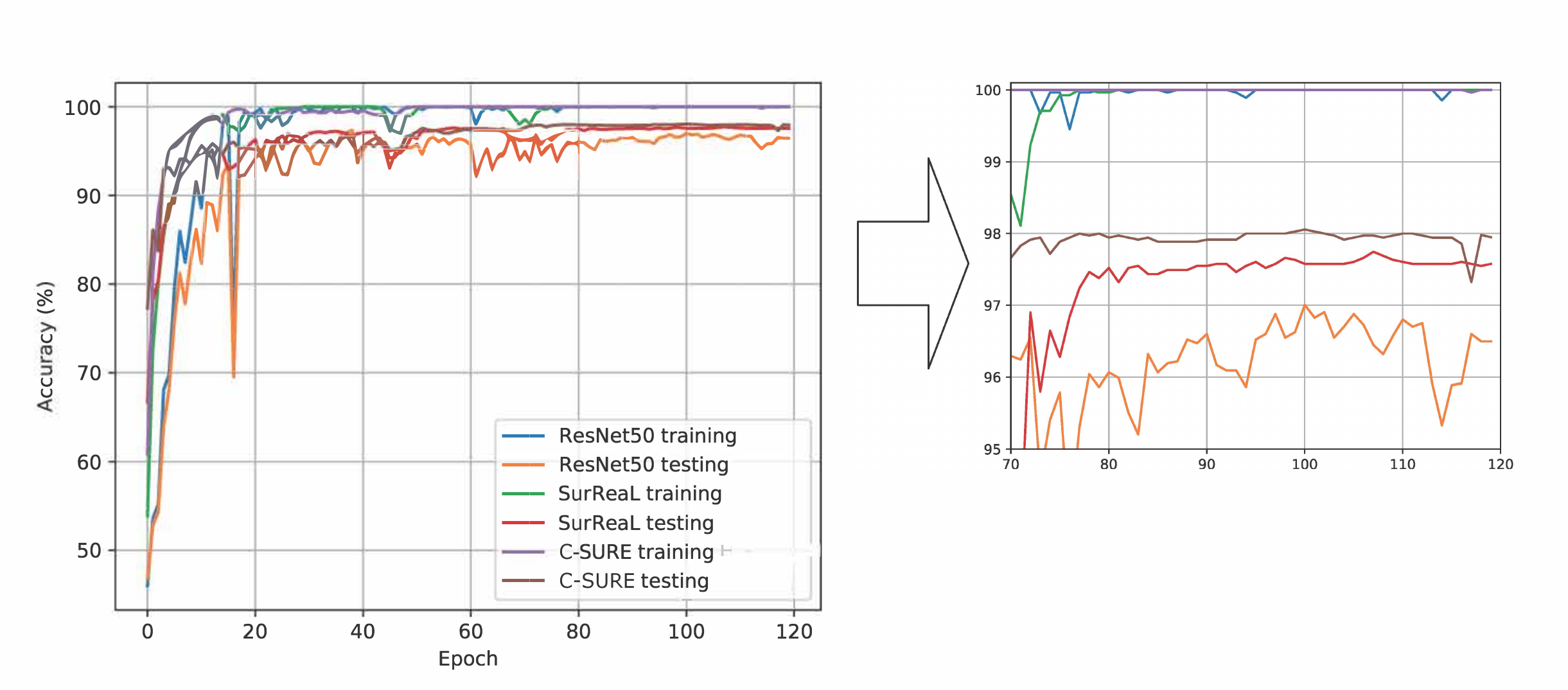}
    \caption{C-SURE is more robust, stable and fast converging.  Shown here is the classification accuracy over training epochs on the MSTAR small set, for ResNet50, SurReal, and C-Sure.  All three models have a significant performance gap between training and testing.  However,  C-SURE has the least gap and is more robust. C-SURE is also more stable and fast converging, as the training accuracy plateaus sooner.}
    \label{fig:Seen}
\end{figure}
}

\def\figRadioSNR#1{
\begin{figure}[#1]
    \centering
    \includegraphics[width=0.45\textwidth]{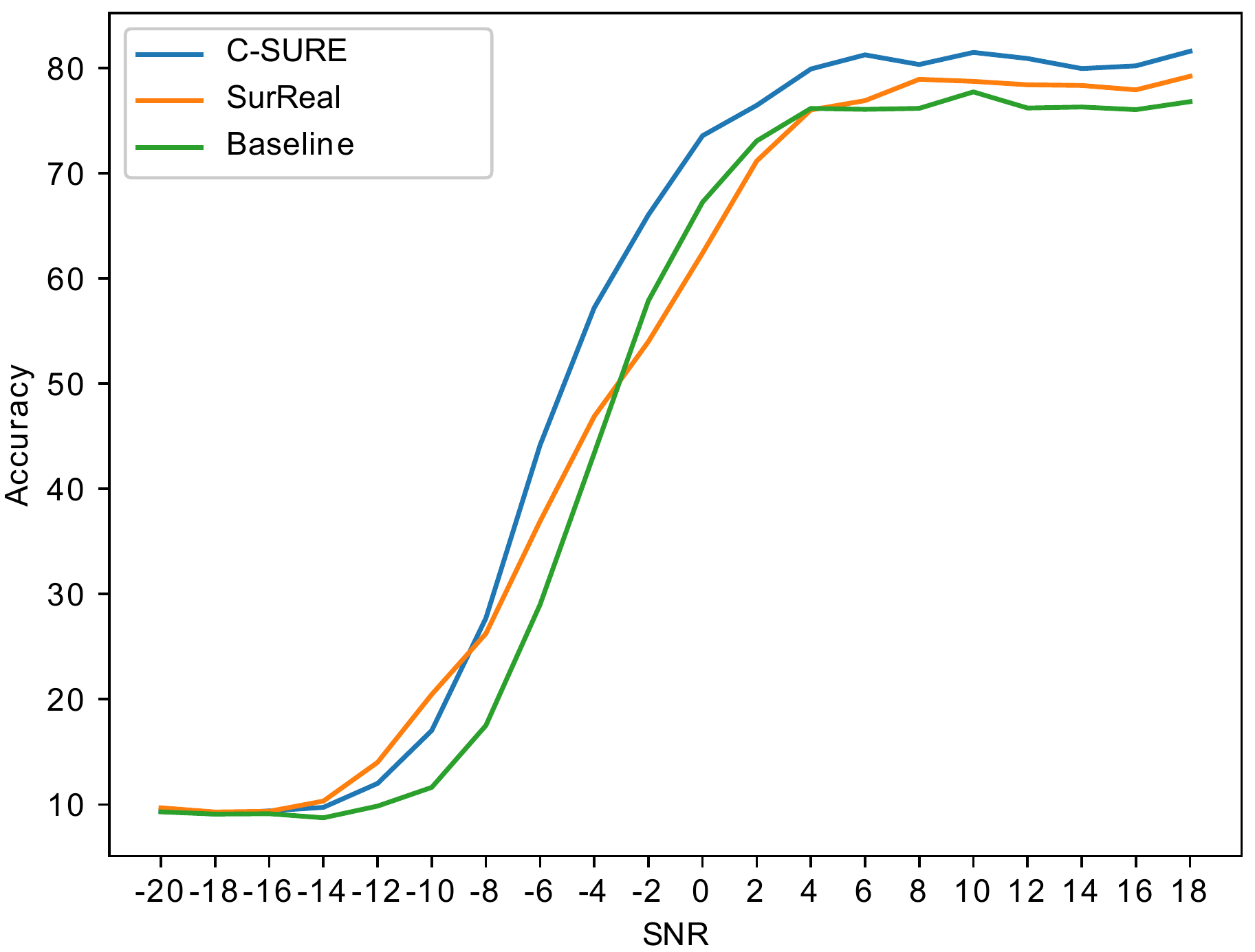}
    \caption{C-SURE has a higher test accuracy than baselines overall.  
      C-SURE outperforms the real-valued baseline at every SNR;  C-SURE outperforms SurReal when SNR$>\!-8$, and the gain is larger when SNR$\in[-8,8]$.  }
    \label{fig:SNR}
\end{figure}
}

\def\tabV#1{
\begin{table}[#1]
\centering
\caption{C-SURE with Varying Hyperparameter $v$}
\label{tabv}
\setlength{\tabcolsep}{10pt}
\begin{tabular}{l|r|c}
\toprule

\textbf{Dataset} & \bf Variance $v$ & \textbf{Accuracy} (\%) \\
\midrule

\multirow{3}{4.5em}{\textbf{MSTAR}} & 
(MLE) \hspace{8pt} 0 & 98.8 \\
\cline{2-3}
& 1 & \bf 99.2 \\
\cline{2-3}
& 10 & 97.5 \\

\midrule

\multirow{3}{4.5em}{\textbf{RadioML}} & 
(MLE)\hspace{8pt} 0 & 80.7 \\
\cline{2-3}
& 1 & \bf 81.6 \\
\cline{2-3}
& 10 & 78.6 \\

\bottomrule
\end{tabular}
\end{table}
}

\section{Experimental Results}

We compare our C-SURE prototype classifier
against two baselines.  {\bf 1)} The first baseline is a real-valued CNN classifier which ignores the geometry of complex numbers and treats each complex value as two independent real numbers. 
{\bf 2)} The second baseline is the complex-valued SurReal discriminative classifer our model is built on.  

We experiment on two complex-valued datasets: MSTAR and RadioML.  Our results demonstrate that our model-based classifier is more accurate (Table \ref{resultSummary}), more stable and robust  (Fig. \ref{fig:Seen}).   Like SurReal, our model is also much smaller, outperforming the real-valued CNN on MSTAR with less than $1\%$ of the model size (Fig. \ref{fig:mstar_params}).

\figResSum{!hp}
\figModelSize{!hp}

All the experiments are trained on a GeForce RTX $2080$ GPU for a total of $120$ epochs, using Adam optimizer and cross-entropy loss. The batch size is $100$ for MSTAR and $400$ for RadioML.  The learning rate is $0.015$ for MSTAR and $0.03$ for RadioML.

\subsection{MSTAR Target Classification}



\noindent
{\bf  MSTAR Data.}
The dataset contains complex-valued SAR images of $11$ classes \cite{keydel1996mstar}.
We create two random subsets, large (L) and small (S), from the original MSTAR dataset.  The small set is entirely contained in the large set.
We center-crop the SAR images into $100\times 100$ pixels, and convert the complex values into the polar form.
\\

\figMSTARSets{hp}
\figAngleConf{tp}
\figSeen{tp}

\noindent
{\bf  Real-valued CNN Baseline.}
We use ResNet50 \cite{he2016deep} and feed the complex-valued image as a (real,imaginary) two-channel real-valued image.
Fig. \ref{fig:mstar_params} compares the model size among different approaches and implementations.  Even with the need to store the class prototypes, our C-SURE classifier remains light-weight like SurReal,  with less than $1\%$ of the ResNet50 size.
\\

\noindent
{\bf Accuracy and Robustness.}
Table \ref{resultSummary} shows that C-SURE is overall more accurate than SurReal and  ResNet50, and the gain is larger for the small dataset, with least confusion between classes (Fig. \ref{fig:conf}).  This slight effect is consistent with the idea of using prototypes for few-shot recognition.

Fig. \ref{fig:Seen} compares how the training and testing accuracy evolves during training.  C-SURE seems not only more stable and fast converging as the training accuracy plateaus sooner, but also more robust as 
it has the least performance gap between training and testing.

\subsection{RadioML Modulation  Classification}

\figRadioSNR{tp}

\noindent
{\bf RadioML Data.}
They are synthetically generated radio signals with modulation operating over both voice and text data.  Noise is added further for channel effects.  
Each signal is tagged with a signal-to-noise ratio (SNR), in the range of $[-20,18]$ with an increment step of $2$.  There are $11$ types of modulations; each type has $20,000$ instances.   The data is split $50/50$ between training and testing. 
\\

\noindent
{\bf  Real-valued CNN Baseline.}
We use the \'OShea's model \cite{o2016convolutional} and feed the complex-valued RF time series of length 128 as a (real,imaginary) two-channel signal.
Our C-SURE classifier is less than $3\%$ of the size of \'OShea's model. 
\\

\noindent
{\bf Accuracy over SNR.}
Fig. \ref{fig:SNR} compares the test accuracy at various SNR's.
When the SNR is too low or too high, all three models become almost equally poor or good.
Nevertheless, C-SURE is more accurate than the real-valued baseline at every SNR, than  SurReal when SNR$>\!-8$, with a larger gain in the middle SNR range of $[-8,8]$.

\subsection{Shrinkage or MLE in A Prototype Classifier?}

Our C-SURE classifier differs from SurReal, the complex-valued classifier baseline, on two aspects: It models class prototypes explicitly and uses the  shrinkage estimator on the manifold of complex values.  If we fix the model as a prototype classifier and vary the amount of shrinkage,  we can tease out the contribution of the shrinkage estimator against the standard MLE in our C-SURE classifier.

\tabV{tp}

The C-SURE shrinkage estimator has a hyperparameter $v$ specifying the data variance in the hierarchical Bayesian model.  When $v=0$, there is no shrinkage adjustment from the prior distribution, and the estimator is reduced to MLE.

Table \ref{tabv} lists the test accuracies on both MSTAR and RadioML tasks as we vary the hyperparameter $v$.  There is always a shrinkage estimator at $v>0$ better than MLE -- the shrinkage estimator at $v=0$, validating the benefit of utilizing a shrinkage estimator in a prototype CNN classifier.

While C-SURE seems to be able to outperform SurReal, the size of gain remains small.  More controlled and careful experimentation would be needed to clarify which data classification scenarios C-SURE would be best at.

\section{Summary}
\label{conc}

Most existing deep learning approaches assume  data lying in a vector space.  We consider the complex-valued data, where the range of the data is no longer in the Euclidean space.  The SurReal complex-valued classifier outperforms the real-valued CNN baseline with a significantly reduced model size \cite{chakraborty2019real}, based on  computing the geometric mean on the manifold (i.e., FM) for complex-valued data.  

We propose C-SURE, a novel shrinkage estimator on the complex-valued manifold with provably smaller MSE than FM.   We further incorporate it into learning a nearest prototype CNN classifier, by adapting SurReal's model architecture and with only a slight increase in the model size.

On complex-valued MSTAR and RadioML datasets,  our experimental results suggest that our C-SURE classifier tends to be more accurate and robust than SurReal, and the shrinkage estimator is always better than MLE for the same prototype classifier.  

More  experimentation would help clarify the strengths and weaknesses of the C-SURE classification approach.

\section*{Acknowledgements}
This research was supported, in part, by  Berkeley Deep Drive and DARPA.
{\small
\bibliographystyle{ieee_fullname}
\bibliography{Figures/references}
}
\end{document}